\documentclass[a4paper,UKenglish,cleveref, autoref, thm-restate,authorcolumns]{lipics-v2019}
\usepackage{wrapfig}
\usepackage{centernot}
\usepackage{paralist}
\usepackage{tikz}
\usepackage{graphicx}
\usepackage{xspace}
\usepackage{amssymb}
\usepackage{amsmath}
\usepackage{paralist}
\usepackage{mathtools}

\newcommand\lorplus{\mathbin{\ooalign{$\vee$\cr%
   \hfil\raise0.65ex\hbox{$\scriptscriptstyle +$}\hfil\cr}}}

\newcommand{\finishes}{\ensuremath{\langle E \rangle}\xspace}
\newcommand{\during}{\ensuremath{\langle D \rangle}\xspace}
\newcommand{\starts}{\ensuremath{\langle B \rangle}\xspace}
\newcommand{\overlaps}{\ensuremath{\langle O \rangle}\xspace}
\newcommand{\meets}{\ensuremath{\langle A \rangle}\xspace}

\newcommand{\later}{\ensuremath{\langle L \rangle}\xspace}

\newcommand{\mmodels}{\ensuremath{\Vdash}}

\title{Knowledge Extraction with Interval Temporal Logic Decision Trees} 

\author{Guido Sciavicco}{Department of Mathematics and Computer Science\\{University of Ferrara, Italy}}{guido.sciavicco@unife.it}{http://orcid.org/0000-0002-9221-879X}{}

\author{Ionel Eduard Stan}{Department of Mathematics and Computer Science\\{University of Ferrara, Italy} \\ {\&} \\{Department of Mathematical, Physical, and Computer Sciences} \\ {University of Parma, Italy}}{ioneleduard.stan@unife.it}{http://orcid.org/0000-0001-9260-102X}{}

\authorrunning{G. Sciavicco and I.E. Stan} 

\Copyright{G. Sciavicco and I.E. Stan} 

\ccsdesc[500]{Theory of computation}
\ccsdesc[500]{Theory of computation~Logic}

\keywords{Interval Temporal Logic, Decision Trees, Explainable AI, Time series.} 

\acknowledgements{Computational resources have been offered by the University of Udine, Italy, supported by the PRID project
\emph{Efforts in the uNderstanding of Complex interActing SystEms} (ENCASE) and the authors acknowledge the partial support by the Italian INDAM~GNCS project \emph{Strategic Reasoning and Automated Synthesis of Multi-Agent Systems}}

\nolinenumbers 

\EventEditors{Emilio Mu\~{n}oz-Velasco, Ana Ozaki, and Martin Theobald}
\EventNoEds{3}
\EventLongTitle{27th International Symposium on Temporal Representation and Reasoning (TIME 2020)}
\EventShortTitle{TIME 2020}
\EventAcronym{TIME}
\EventYear{2020}
\EventDate{September 23--25, 2020}
\EventLocation{Bozen-Bolzano, Italy}
\EventLogo{}
\SeriesVolume{178}
\ArticleNo{6}

\begin{document}

\maketitle

\begin{abstract}
Multivariate temporal, or time, series classification is, in a way, the temporal generalization of (numeric) classification, as every instance is described by multiple time series instead of multiple values. Symbolic classification is the machine learning strategy to extract explicit knowledge from a data set, and the problem of symbolic classification of multivariate temporal series requires the design, implementation, and test of ad-hoc machine learning algorithms, such as, for example, algorithms for the extraction of temporal versions of decision trees. One of the most well-known algorithms for decision tree extraction from categorical data is Quinlan's ID3, which was later extended to deal with numerical attributes, resulting in an algorithm known as C4.5, and implemented in many open-sources data mining libraries, including the so-called Weka, which features an implementation of C4.5 called J48. ID3 was recently generalized to deal with temporal data in form of timelines, which can be seen as discrete (categorical) versions of multivariate time series, and such a generalization, based on the interval temporal logic HS, is known as Temporal ID3. In this paper we introduce Temporal C4.5, that allows the extraction of temporal decision trees from undiscretized multivariate time series, describe its implementation, called Temporal J48, and discuss the outcome of a set of experiments with the latter on a collection of public data sets, comparing the results with those obtained by other, classical, multivariate time series classification methods.
\end{abstract}

\section{Introduction}

A {\em labelled data set} $\mathcal D=\{D_1,\ldots,D_m\}$ is a set of instances described by a set of numerical and/or categorical attributes $\mathcal A=\{A_1,\ldots,A_n\}$ and associated to a set of classes $\mathcal C=\{C_1,\ldots,C_q\}$. {\em Supervised symbolic classification} is the machine learning strategy for extracting an explicit (logical) theory that describes a labelled data set $\mathcal D$. It is usually opposed to {\em supervised functional classification}, which includes linear regression, logistic regression, neural networks, and many other black-box and function-extraction mechanisms. There are many symbolic classification methods, which can be broadly distinguished into {\em tree-based} and {\em rule-based}. Tree-based classification models can be {\em single} or {\em multiple}; multiple tree-based models, however, while still symbolic in nature, are not interpretable in the same sense as single trees are. Tree-based classification models are also known as {\em decision trees}, and they can be described in a very abstract (inductive) way: a decision tree is a class, or it is a $k$-ways decision followed by $k$ decision trees. The introduction of decision trees can be dated back to~\cite{DBLP:journals/ml/Quinlan86,Quinlan93}. The problem of extracting the {\em optimal} decision tree from a data set is NP-hard~\cite{DBLP:journals/ipl/HyafilR76}, which justifies the use of sub-optimal approaches. The {\em ID3} algorithm~\cite{DBLP:journals/ml/Quinlan86} is a greedy approach to decision tree extraction; it is based on a simple concept: given a labelled data set $\mathcal D$, one takes a {\em decision} by choosing the attribute $A_i$ and the value(s) on the domain $A_i$ that return(s) the highest information, obtaining, in general, a $k$-ways partition of $\mathcal D$ in $\mathcal D_1,\ldots,\mathcal D_k$. The information is measured in terms of the classes that occur in $\mathcal D$ and in $\mathcal D_1,\ldots,\mathcal D_k$. Each decision can be expressed as a propositional letter; since alternative branches can be seen as logical (exclusive) disjunctions, and successive decisions on the same branches can be seen as logical conjunctions, a decision tree as a whole can be seen as a set of propositional formulas. In other words, a decision tree is a propositional description of the data set on which it is learned.

\medskip

A {\em time series} is a set of variables that change over time, and they can be {\em univariate} or {\em multivariate}. Each variable of a multivariate time series is an ordered collection of $N$ real values, instead of a single value. So, a {\em labelled temporal data set} $\mathcal T=\{T_1,\ldots,T_m\}$ is a set of temporal instances described by a set of temporal attributes $\mathcal A=\{A_1,\ldots,A_n\}$, each being a $N$-points time series, and associated to a set of classes $\mathcal C=\{C_1,\ldots,C_q\}$. Multivariate time (or temporal) series emerge in many application contexts. The temporal history of some hospitalized patient can be described by the time series of the values of his/her temperature, blood pressure, and oxygenation; the pronunciation of a word in sign language can be described by the time series of the relative and absolute positions of the ten fingers w.r.t. some reference point; different sport activities can be distinguished by the time series of some relevant physical quantities. In the current literature, time series classification algorithms can be instance-based, feature-based, and timeline-based. {\em Instance-based} methods are essentially all built on the notion of distance between two time series, by means of which a time series can be classified using, e.g.,  the {\em Nearest Neighbor} (NN) algorithm. The most widely accepted notions of distances are the {\em Euclidean Distance} (ED) and the {\em Dynamic Time Warping} (DTW)~\cite{SWK2015}. Intuitively, ED is a one-to-one alignment method, while DTW is one-to-many, as it allows one to compare time series even of different scales. Such methods have been systematically applied to a variety of multivariate time series data sets in~\cite{bagnall2018uea} (the univariate case is dealt with by the same authors in~\cite{DBLP:journals/datamine/BagnallLBLK17}). {\em Feature-based} methods, on the other hand, consist of flattening the time series, and describe each one of them via a set of values (e.g., mean, variance, maximum, minimum). These descriptions, in turn, can be used as the input of a static learning algorithm. Feature-based techniques are widespread in the data science community, because they are conceptually simple, and allow one to use familiar learning methods; unfortunately, the theories obtained in this way are not always interpretable, and the quality of the models in term of performances is not always acceptable. An extensive comparison between instance-based and feature-based methods can be found in~\cite{DBLP:journals/tkde/FulcherJ14}, in which the authors also present an algorithm that allows one to automatically choose the best features to represent a time series data set for it to be classified. Finally, A {\em timeline} can be considered as the discretized version of a multivariate time series. For each single variable, one produces a set of propositional letters that describe the values of that variable, or its (first, second, \ldots) derivative, on every possible interval of time, and then describes a multivariate time series on a single line by joining the propositional, interval description of all variables. A general method to translate a multivariate time series into a timeline is described in~\cite{DBLP:conf/iwinac/SciaviccoSV19}, and {\em Temporal ID3}~\cite{DBLP:conf/jelia/BrunelloSS19} can be considered an example of {\em timeline-based} classification of multivariate time series, that uses the interval temporal logic HS~\cite{HalpernS91} to describe a decision tree.

\medskip

In this paper we design a decision tree learning algorithm, {\em Temporal C4.5}, that generalizes Temporal ID3 in the same way in which the algorithm C4.5~\cite{Quinlan:1987:SDT:50007.50008} generalizes ID3, that is, by introducing the possibility of learning from continuous attributes. In this case, however, (non-discretized) time series are described {\em only} by continuous (time-changing) values, so, in the current version, Temporal C4.5 does not admit categorical attributes. We consider one of the most representative implementations of C4.5, called {\em J48}, available in the Weka open-source learning suite~\cite{weka}, and we modify it by introducing decisions based on the interval temporal logic HS. The main contributions of this paper are: \begin{inparaenum}[\it (i)]\item the definition of a general theory of decision trees, which is used to guide our generalization from the static to the temporal case; \item the first implementation of a symbolic classification algorithm for time series that deals with the raw data (i.e., without applying any pre-abstraction method), whose extracted theory is expressed in the interval temporal logic HS; \item a comparison of the performances of our implementation against existing methods on the public data used in~\cite{bagnall2018uea}. \end{inparaenum}

%

\section{Preliminaries}\label{sect:preliminaries}



\noindent{\bf Classification of multivariate temporal series.} A {\em time series} is a set of variables that change over time, and they can be {\em univariate} or {\em multivariate}. Each variable (or {\em channel}) of a multivariate time series is an ordered collection of $N$ real values, instead of a single value, so that a single time series can be described as follows:

\begin{equation}\label{eq:multivatiate_time_series}
T=\left\{
\begin{array}{llllll}
A_1 &=& a_{1,1},&a_{1,2},&\ldots,&a_{1,N}\\
A_2 &=& a_{2,1},&a_{2,2},&\ldots,&a_{2,N}\\
\ldots& &\ldots \\
A_n &=& a_{n,1},&a_{n,2},&\ldots,&a_{n,N}.
\end{array}\right.
\end{equation}

\noindent So, a {\em labelled temporal data set} $\mathcal T=\{T_1,\ldots,T_m\}$ is a set of temporal instances described by a set of temporal attributes $\mathcal A=\{A_1,\ldots,A_n\}$, each being a $N$-points time series, and associated to a set of classes $\mathcal C=\{C_1,\ldots,C_q\}$. A temporal data set can be viewed as a $m \times n$ matrix where the $i$th row, $1 \leq i \leq m$, is a multivariate time series and the $j$th column, $1 \leq j \leq n$, is an attribute. The {\em multivariate time series supervised classification problem} is the problem of finding a formula (symbolic classification) or a function (functional classification) that associates multivariate time series to classes.

\begin{figure*}[t]
\centering
\begin{tikzpicture}[scale=.75]
 \tikzstyle{every node}=[font=\footnotesize]

\draw (0,0)node(op){\bf HS};

%

\draw (op) ++(0,-1.5)node(meets){$\meets$};
\draw (meets)++(0,-.75)node(later){$\later$};
\draw (meets)++(0,-1.5)node(starts){$\starts$};
\draw (meets)++(0,-2.25)node(finishes){$\finishes$};
\draw (meets)++(0,-3)node(during){$\during$};
\draw (meets)++(0,-3.75)node(overlaps){$\overlaps$};


\draw (meets)++(1.7,0) node(sep1){} ++(0,.5) -- ++(0,-4.7);


\draw (sep1)++(3,1.5)node(rel){\bf Allen's relations};

\draw (sep1)++(0.5,0)node[right](Ra){$[x,y] R_A [x',y'] \Leftrightarrow y=x'$};
\draw (sep1)++(0.5,-.75)node[right](Rl){$[x,y] R_L [x',y'] \Leftrightarrow y < x'$};
\draw (sep1)++(0.5,-1.5)node[right](Rs){$[x,y] R_B [x',y'] \Leftrightarrow x=x', y' < y$};
\draw (sep1)++(0.5,-2.25)node[right](Rf){$[x,y] R_E [x',y'] \Leftrightarrow y=y', x < x'$};
\draw (sep1)++(0.5,-3)node[right](Rd){$[x,y] R_D [x',y'] \Leftrightarrow x < x', y' < y$};
\draw (sep1)++(0.5,-3.75)node[right](Ro){$[x,y] R_O [x',y'] \Leftrightarrow x < x' < y < y'$};


\draw (sep1)++(8.5,0)node(sep2){} ++(0,0.5) -- ++(0,-4.7);

\draw (sep2)++(2.8,1.5)node(graphic){\bf Graphical representation};

\draw[red,|-|] (sep2) ++(.7,.75)node[above](a){\tiny $x$} -- ++(2,0)node[above](b){\tiny $y$};
\draw[dashed,red,help lines,thick] (a) -- ++(0,-5.25);
\draw[dashed,red,help lines,thick] (b) -- ++(0,-5.25);

\draw[|-|] (b) ++(0,-1) ++(0,0)node[above](Ac){\tiny $x'$}
-- ++(1,0)node[above](Ad){\tiny $y'$};
\draw[|-|] (b) ++(0,-1) ++(.5,-.75)node[above](Lc){\tiny $x'$}
-- ++(1,0)node[above](Ld){\tiny $y'$};
\draw[|-|] (a) ++(0,-1) ++(0,-1.5)node[above](Bc){\tiny $x'$}
-- ++(.5,0)node[above](Bd){\tiny $y'$};
\draw[|-|] (b) ++(0,-1) ++(-.5,-2.25)node[above](Ec){\tiny $x'$}
-- ++(.5,0)node[above](Ed){\tiny $y'$};
\draw[|-|] (a) ++(0,-1) ++(.5,-3)node[above](Dc){\tiny $x'$}
-- ++(1,0)node[above](Dd){\tiny $y'$};
\draw[|-|] (a) ++(0,-1) ++(1,-3.75)node[above](Oc){\tiny $x'$}
-- ++(2,0)node[above](Od){\tiny $y'$};
\end{tikzpicture}


\caption{Allen's interval relations and HS modalities.}
\label{fig:relations}
\end{figure*}

\medskip

\noindent {\bf Timelines and interval temporal logic}. A multivariate time series can be discretized without eliminating the temporal component. In~\cite{DBLP:conf/iwinac/SciaviccoSV19} the authors introduce the notion of {\em timeline} and present a procedure that transforms multivariate time series into timelines. Time series describe continuous processes; when discretized, it makes little sense to model their values at each point, but, instead, they are naturally represented in a interval-based ontology. Thus, if a static numerical data set is naturally represented in propositional logic, a multivariate time series is naturally represented in an interval temporal logic.

\medskip
Let $[N]$ an initial subset of $\mathbb N$ of length $N$. An {\em interval} over $[N]$ is an ordered pair $[x,y]$, where $x,y \in [N]$ and $x < y$, and we denote by $\mathbb I(\mathbb [N])$ the set of all intervals over $[N]$. If we exclude the identity relation, there are 12 different Allen's relations between two intervals in a linear order~\cite{allen83}: the six relations $R_A$
(adjacent to), $R_L$ (later than), $R_B$ (begins), $R_E$ (ends), $R_D$ (during), and
$R_O$ (overlaps), depicted in Figure~\ref{fig:relations}, and their inverses, that is, $R_{\bar{X}}
= (R_{X})^{-1}$, for each $X \in \mathcal X$, where $\mathcal X=\{ A, L, B, E, D, O \}$. Halpern and Shoham's modal logic of temporal intervals (HS) is defined from a set of propositional letters $\mathcal{AP}$, and by associating a universal modality $[X]$ and an existential one $\langle X\rangle$ to each Allen's relation $R_{X}$. Formulas of HS are obtained by

\begin{equation}\label{eq:grammar_hs}
\varphi ::= p \mid \neg \varphi \mid \varphi \vee \varphi \mid \langle X \rangle
\varphi\mid \langle \bar X \rangle
\varphi,
\end{equation}

\noindent where $p \in \mathcal{AP}$ and $X \in \mathcal X$. The other Boolean connectives and the logical constants, e.g., $\rightarrow$ and $\top$, as well as the universal modalities $[X]$, can be defined in the standard way, i.e., $[X]p\equiv\neg \langle X\rangle \neg p$. For each $X \in \mathcal X$, the modality $\langle \bar{X} \rangle$ (corresponding to the inverse relation $R_{\bar{X}}$ of $R_{X}$) is said to be the {\em transpose} of the modalities $\langle X \rangle$, and vice versa. The semantics of HS formulas is given in terms of {\em timelines} $T
= \langle \mathbb I([N]),V\rangle$\footnote{We deliberately use the symbol $T$ to indicate both a timeline and a time series.}, where $V : \mathcal{AP} \rightarrow 2^{\mathbb{I}([N])}$ is a \emph{valuation function}
which assigns to each atomic proposition $p \in \mathcal{AP}$ the set of intervals $V(p)$
on which $p$ holds. The {\em truth} of a formula $\varphi$ on a given interval $[x,y]$ in an interval model $T$ is defined by structural induction on formulas as follows:

\begin{equation}\label{eq:semantics_hs}
\begin{array}{lll}
T,[x,y]\mmodels p& \mbox{if}&[x,y]\in V(p),\mbox{ for }p\in\mathcal{AP};\\
T,[x,y]\mmodels\neg\psi&\mbox{if}&T,[x,y]\not\mmodels\psi;\\
T,[x,y]\mmodels\psi\vee\xi&\mbox{if}&T,[x,y]\mmodels\psi\mbox{ or }T,[x,y]\mmodels\xi;\\
T,[x,y]\mmodels\langle X\rangle\psi&\mbox{if}& \mbox{there is }[w,z]\mbox{ s.t }[x,y] R_{X} [w,z]\mbox{ and }T,[w,z]\mmodels\psi;\\
T,[x,y]\mmodels\langle \bar X\rangle\psi&\mbox{if}& \mbox{there is }[w,z]\mbox{ s.t }[x,y] R_{\bar X} [w,z]\mbox{ and }T,[w,z]\mmodels\psi.
\end{array}
\end{equation}

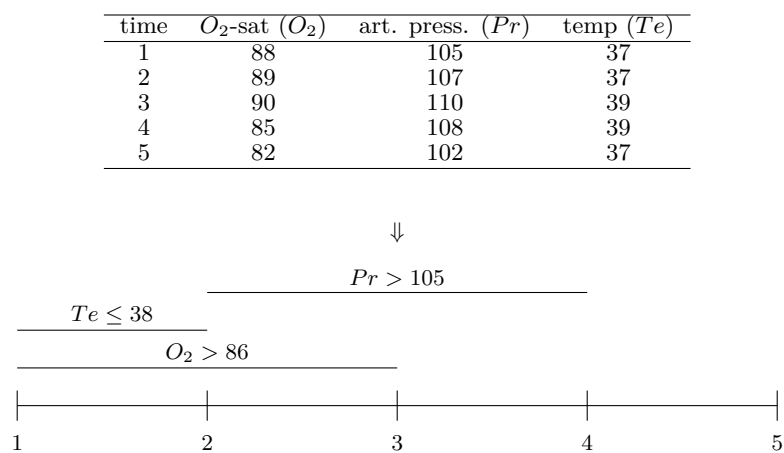
\begin{figure}[t]
\centering\footnotesize
\begin{tabular}{cccc}
\hline
time & $O_2$-sat ($O_2$) & art. press. ($Pr$) & temp ($Te$)\\
\hline
1& 88 & 105 & 37 \\
2& 89 & 107 & 37 \\
3& 90 & 110 & 39 \\
4& 85 & 108 & 39 \\
5& 82 & 102 & 37 \\
\hline
\end{tabular}

\begin{tikzpicture}
\node(void) at (0,3){   };
\node(g) at (5,2.3){$\Downarrow$};
\draw(0,0) -- (10,0);
\draw(0,0.2) -- (0,-0.2);
\draw(2.5,0.2) -- (2.5,-0.2);
\draw(5,0.2) -- (5,-0.2);
\draw(7.5,0.2) -- (7.5,-0.2);
\draw(10,0.2) -- (10,-0.2);
\draw(0,0.5) -- (5,0.5);
\draw(0,1) -- (2.5,1);
\draw(2.5,1.5) -- (7.5,1.5);
\node(void) at (0,-1.5){   };
\node(p0) at (0,-0.5) {$1$};
\node(p1) at (2.5,-0.5) {$2$};
\node(p2) at (5,-0.5) {$3$};
\node(p3) at (7.5,-0.5) {$4$};
\node(p4) at (10,-0.5) {$5$};
\node(o) at (2.5,0.7) {$O_2> 86$};
\node(t) at (1.25,1.2) {$Te\le 38$};
\node(p) at (5,1.7) {$Pr> 105$};
\end{tikzpicture}
\caption{Using timelines to discretize multivariate time series: an example with a single time series.}\label{fig:disc}
\end{figure}


A time series can be seen as a timeline based on a suitable propositional signature. As for example, consider a time series that records medical values of some hospitalized patient: temperature, blood pressure, and oxygenation, as in Figure~\ref{fig:disc}, top. The information can be adequately subsumed into a timeline such as in Figure~\ref{fig:disc}, bottom, provided that a suitable propositional signature is given. In our example, we (arbitrarily) decided that, for instance, the value $38$ is an informative splitting point, and a propositional letter $(Te\le 38)$ can be created. In~\cite{DBLP:conf/iwinac/SciaviccoSV19}, a methodology that allows one to perform such discretization is presented, and in~\cite{DBLP:conf/jelia/BrunelloSS19}, a temporal decision tree extraction method (Temporal ID3), that takes a temporal data set in form of timelines and extracts a decision tree whose edges are labelled with decisions written in HS, is studied. The main limitation of such an approach is that the discretization method does not take into account the predictive capabilities of the decisions (that is, of the propositional letters), because it is run off-line, so to say: in machine learning terms, it is a {\em filter}. Temporal C4.5 is aimed to close precisely this gap, allowing the propositional signature to emerge {\em during} the decision tree extraction, not before, precisely as C4.5 does on non-temporal data.

\section{A Theory of Decision Trees}\label{sec:theory}

\noindent\textbf{A theory of static decision trees.} Decision trees are a very well-known construct. While in the literature there has been a great effort to improve their implementation, versatility, and applicability, a formal definition of the structure and semantics of decision trees is necessary to correctly define their temporal generalization. In this section, for the sake of simplicity of explanation, we restrict our attention to the case of binary decision trees for binary classification, both in the static and the temporal case. Generalizing our approach to the case of $k$-ary trees and multiple classes is immediate.

\medskip

Consider a labelled static data set $\mathcal D=\{D_1,\ldots,D_m\}$ described by the set of attributes $\mathcal A=\{A_1,\ldots,A_n\}$ and associated to the classes $\mathcal C=\{Yes,No\}$. We denote the {\em domain} of an attribute $A$ by $dom(A)$. The language of static decision trees encompasses a set $\mathcal S$ of {\em propositional decisions}:

\begin{equation}\label{eq:decisions_static_dt}
\mathcal{S} = \{A \bowtie a \mid A \in \mathcal{A},a \in dom(A)\},
\end{equation}

\noindent where $\bowtie \, \in \{\le, =\}$. Binary static decision trees are formulas of the following grammar:


\begin{equation}\label{eq:syntax_static_dt_rewritten}
\begin{split}
\hat{\varphi} &::= (S \land \hat{\varphi}) \lorplus (\neg S \land \hat{\varphi}) \mid  C. \\
\end{split}
\end{equation}

\noindent where $C \in \mathcal{C}$ and $S\in\mathcal S$. The symbol $\lorplus$ indicates the exclusive {\em or}, while the symbol $\wedge$ indicates the classical propositional {\em and}. Every non-leaf node of a decision tree has two children and every edge is decorated with a decision. Leaves are decorated with a class. A decision $S$ is interpreted over a single instance $D$ using classical propositional logic. We say that $D$ {\em satisfies} the decision $A \le a$ (resp., $A=a$) if the attribute $A$ has a value less than or equal to (resp., equal to) $a\in dom(A)$ in $D$, and we use the symbol $D\models (A \le a)$ (resp., $D\models (A<a)$).

\medskip

A decision tree is interpreted over a labelled data set $\mathcal D$ via the semantic relation $\hat{\models}_\theta$, which generalizes $\models$ from single instances to data sets: we need to define the notion of a data set satisfying $\hat\varphi$ with parameter $\theta$, that is, $\mathcal D\hat\models_\theta\hat\varphi$. Unlike the classical propositional logic, in this case the truth relation is parametric; the parameter $\theta$ formalizes the notion of how well a decision tree $\hat\varphi$ represents $\mathcal D$. Let $D$ an instance of $\mathcal D$. We denote by $C(D)$ the true class of $D$, and by $\hat\varphi(D)$ the class that $\hat\varphi$ predicts for $D$. The generic performance of $\hat\varphi$ on $\mathcal D$ can be measured in terms of its {\em confusion matrix}, which, for each given instance, expresses one of four possible, mutually exclusive, indicators ({\em true positive}, {\em true negative}, {\em false positive}, and {\em false negative}) by comparing $C(D)$ and $\hat\varphi(D)$:

\begin{equation}
\begin{tabular}{ccc}
                            & $C(D) = No$                                        & $C(D) = Yes$                                        \\ \cline{2-3}
\multicolumn{1}{c|}{$\hat{\varphi}(D) = No$} & \multicolumn{1}{c|}{True Negative (TN)}  & \multicolumn{1}{c|}{False Negative (FN)} \\ \cline{2-3}
\multicolumn{1}{c|}{$\hat{\varphi}(D) = Yes$} & \multicolumn{1}{c|}{False Positive (FP)} & \multicolumn{1}{c|}{True Positive (TP)}  \\ \cline{2-3}
\end{tabular}
\end{equation}

\noindent The root of a tree $\hat\varphi$ is associated with the data set $\mathcal D$ on which it is interpreted, and, in general, each node of the tree is associated with a subset $\mathcal D'\subset\mathcal D$ and a binary decision $S$. A set $\mathcal D'$ is partitioned into two subsets $\mathcal D'_{1}$ and $\mathcal D'_{2}$, that contain, respectively the instances that satisfy $S$ and those that do not. The subset of $\mathcal D$ associated with a leaf is also labelled with a class $C$, meaning that every instance in it is classified with $C$, generating a certain amount of misclassifications. From the leaves, one can inductively compute the confusion matrix of each node. The confusion matrix of the root is the one we associate with the tree itself. The rules for $\hat\models_\theta$ are now immediate:

\begin{equation}\label{eq:sematincs_static_dt}
\begin{tabular}{lcl}
$\mathcal D \, \hat{\models}_\theta No$ & if & $\theta =$
			\begin{tabular}{|c|c|}
                \hline
                $|\mathcal{D}_{No}|$ & $|{\mathcal{D}}| - |{\mathcal{D}}_{No}|$ \\ \hline
                0 & 0 \\ \hline
            \end{tabular}{}
            , where \\ & & ${\mathcal{D}}_{No} = \{{D} \in {\mathcal{D}} \mid C({D}) = No\}$, \\
${\mathcal{D}} \, {\hat\models}_\theta Yes$ & if & $\theta =$
			\begin{tabular}{|c|c|}
                \hline
                0 & $0$ \\ \hline
                $|{\mathcal{D}}| - |\mathcal{D}_{Yes}|$ & $|{\mathcal{D}}_{Yes}|$ \\ \hline
            \end{tabular}{}
            , where \\ & & ${\mathcal{D}}_{Yes} = \{D \in {\mathcal{D}} \mid C({D}) = Yes\}$, \\
${\mathcal{D}} \, {\hat\models}_\theta (S \land {\hat\varphi}_1) \lorplus (\neg S \land {\hat\varphi}_2)$ & if & $\theta = \theta_1 + \theta_2,{\mathcal{D}}_1 \, {\hat\models}_{\theta_1} {\hat\varphi}_1$, and ${\mathcal{D}}_2 \, {\hat\models}_{\theta_2} {\hat\varphi}_2$, where \\
& & ${\mathcal{D}}_1 = \{{D} \in {\mathcal{D}} \mid {D} \models S \}, {\mathcal{D}}_2 = \{{D} \in {\mathcal{D}} \mid {D} \models \neg S \}$,\\
& & ${\mathcal{D}} = {\mathcal{D}}_1 \cup {\mathcal{D}}_2,$ and ${\mathcal{D}}_1 \cap {\mathcal{D}}_2 = \emptyset.$
\end{tabular}
\end{equation}

\noindent Observe that computing the confusion matrix on a node generalizes every classical notion of performance, such as accuracy, precision, recall, among others.

\medskip

\noindent \textbf{A theory of temporal decision trees.} On the basis of the above notions, we can now define the concept of temporal decision tree.

\medskip

Let us now consider a labelled temporal data set $\mathcal T=\{T_1,\ldots,T_n\}$, in which every instance is described by the time series $\mathcal A=\{A_1,\ldots,A_n\}$ (recall that each attribute is a univariate time series, in this case) and, as before, classified in one of two classes from the set $\mathcal C=\{Yes,No\}$. We assume that all attributes have the same temporal length, $N$. Because time series describe continuous processes, our decisions will be taken on the value of a single channel over an interval of time, and we describe such decisions using propositional interval temporal logic. Unlike static attributes, however, temporal attributes can be analyzed over multiple dimensions. Consider an interval of time $[x,y]$ and an attribute $A$ that varies on it. As in the static case we can ask the question $A\bowtie a$ over the entire interval, where $\bowtie \, \in\{\le,=\}$, which is positively answered if {\em every value} of $A$ in the interval $[x,y]$ respects the given constraint. But unlike the static case, we do not ask if $A\bowtie a$ only in the {\em current interval} but also if {\em there exists an interval}, related to the current one, in which that holds, so that the decision becomes $\langle X\rangle(A\bowtie a)$. This implies, among other things, that the relation $>$ cannot be defined as the negation of $\le$: when we apply the negation, indeed, we negate {\em both} $\langle X\rangle$ and $\bowtie$, which amounts to say that if we want to check if $\langle X\rangle(A>a)$ we have to do it explicitly. Therefore, in this case, $\bowtie \, \in\{\le,=,>\}$. Moreover, in order to allow a certain degree of uncertainty, we may relax the requirement $A\bowtie a$ over a given interval $[x,y]$ by asking that {\em at least a certain fraction} of the values of $A$ in the interval $[x,y]$ meet this condition; we denote this relaxed decision by $A\bowtie_\alpha a$, where $\alpha\in (0,1]$ is a real parameter. Finally, we need to remember that in certain applications the values may not be as important as the trends, that is, the value of the {\em discrete derivative} of $A$ at a certain degree. We denote the discrete derivative of $A$ at degree $z$ by $A^z$ (identifying $A^0$ with $A$), and, consequently, a generic temporal decision by $\langle X\rangle(A^z\bowtie_\alpha a)$, with $a\in dom(A^z)$. Thus, the language of temporal decision trees encompasses the following set of {\em temporal decisions}:

\begin{equation}\label{eq:decisions_temporal_dt}
\begin{array}{lll}
\mathcal{S} & = & \{\langle X \rangle (A^z \bowtie_\alpha a), \langle \bar{X} \rangle (A^z \bowtie_\alpha a) \mid X \in \mathcal{X}, A \in \mathcal{A},a \in dom(A^z)\} \, \cup \\
                  && \{A^z \bowtie_\alpha a \mid A \in \mathcal{A},a \in dom(A^z)\},
\end{array}
\end{equation}

\noindent where $\mathcal{X} = \{A,L,B,E,D,O\}$ are interval operators of the language of HS. Temporal decision trees are formulas obtained from~\eqref{eq:syntax_static_dt_rewritten}, in which propositional decisions have been replaced by temporal decisions. A temporal decision is interpreted over a single multivariate time series $T$ and interval $[x,y]$, by using the notion of semantic relation $\mmodels$ recalled in Section~\ref{sect:preliminaries}; therefore, we use the notation $T,[x,y]\mmodels\langle X\rangle(A^z\bowtie_\alpha a)$ or $T,[x,y]\mmodels\langle \bar X\rangle(A^z\bowtie_\alpha a)$. Formally, given a point $t$, we denote by $A^z(t)$ the value of the $z$-th discrete derivative of the attribute $A$ at the point $t$, and given an interval $[x,y]$, we denote by $[x,y]^{A^z\bowtie a}$ the following set:

\begin{equation}
[x,y]^{A^z\bowtie a} = \{t\mid x\le t\le y,A^z(t)\bowtie a\}.
\end{equation}

\noindent Therefore we have:

\begin{equation}
\begin{array}{lll}
T,[x,y]\mmodels\langle X\rangle (A^z\bowtie_\alpha z) &\mbox{if}& \mbox{there is }[w,z]\mbox{ s.t } [x,y] R_{X} [w,z]\mbox{ and }\\
                                                      && |[w,z]^{A^z\bowtie a}|\ge \lceil\alpha\cdot (z-w+1)\rceil.
\end{array}
\end{equation}

We now want to define the notion of a temporal data set satisfying $\hat\varphi$ with parameter $\theta$, that is, $\mathcal T\hat\mmodels_\theta\hat\varphi$. In the static case, from a data set $\mathcal D$ one computes immediately the two data sets $\mathcal D_1$ and $\mathcal D_2$ entailed by a decision $S$. In the temporal case, however, this requires more effort. Every instance of the temporal data set $\mathcal T$ associated with the root of $\hat\varphi$ is assigned the reference interval $[0,1]$ by default; observe that, since the language of HS encompasses a set of jointly exhaustive operators, this requirement does not decrease the expressive power of temporal decisions. Given a temporal decision $S$ over $\mathcal T$, computing the set $\mathcal T_1$ of all instances that do satisfy $S$ implies assigning to each instance $T\in\mathcal T_1$ a potentially different reference interval; on the contrary, computing $\mathcal T_2$ implies leaving the reference interval of its members unchanged. So, for example, given $\mathcal T$, in which every instance $T$ is assigned a reference interval $[x_T,y_T]$, and given the decision $S=\langle A\rangle(A \le a)$, we say that:

\begin{equation}
\begin{array}{lll}
\mathcal T_1&=&\{ T\in\mathcal T \mid \exists [y_T,z_T] (y_T<z_T\wedge T,[y_T,z_T]\mmodels(A \le a))\}\\
\mathcal T_2&=&\{ T\in\mathcal T \mid \forall [y_T,z_T] (y_T<z_T\rightarrow T,[y_T,z_T]\not\mmodels(A \le a))\}.\\
\end{array}
\end{equation}

\noindent In the particular case in which $S$ is static, the reference interval does not change for $\mathcal T_1$, either. For a temporal decision $S$, we use the notation $T\mmodels S$ (resp., $T\mmodels \neg S$) to identify the members of $\mathcal T_1$ (resp., $\mathcal T_2$). Observe that $S$ entails unique $\mathcal T_1$ and $\mathcal T_2$, but not unique (new) reference intervals for the members of $\mathcal T_1$; however, this choice is implementative, not theoretical. At this point, the notion of how well a temporal data set $\mathcal T$ satisfies a decision tree $\hat\varphi$ becomes immediate, and completely equivalent to the static case:

\begin{equation}\label{eq:sematincs_temporal_dt}
\begin{tabular}{lcl}
$\mathcal T \, \hat{\mmodels}_\theta No$ & if & $\theta =$
			\begin{tabular}{|c|c|}
                \hline
                $|\mathcal{T}_{No}|$ & $|{\mathcal{T}}| - |{\mathcal{T}}_{No}|$ \\ \hline
                0 & 0 \\ \hline
            \end{tabular}{}
            , where \\ & & ${\mathcal{T}}_{No} = \{{T} \in {\mathcal{T}} \mid C({T}) = No\}$, \\
${\mathcal{T}} \, {\hat\mmodels}_\theta Yes$ & if & $\theta =$
			\begin{tabular}{|c|c|}
                \hline
                0 & $0$ \\ \hline
                $|{\mathcal{T}}| - |\mathcal{T}_{Yes}|$ & $|{\mathcal{T}}_{Yes}|$ \\ \hline
            \end{tabular}{}
            , where \\ & & ${\mathcal{T}}_{Yes} = \{T \in {\mathcal{T}} \mid C({T}) = Yes\}$, \\
${\mathcal{T}} \, {\hat\mmodels}_\theta (S \land {\hat\varphi}_1) \lorplus (\neg S \land {\hat\varphi}_2)$ & if & $\theta = \theta_1 + \theta_2,{\mathcal{T}}_1 \, {\hat\mmodels}_{\theta_1} {\hat\varphi}_1$, and ${\mathcal{T}}_2 \, {\hat\mmodels}_{\theta_2} {\hat\varphi}_2$, where \\
& & ${\mathcal{T}}_1 = \{{T} \in {\mathcal{T}} \mid {T} \mmodels S \}, {\mathcal{T}}_2 = \{{T} \in {\mathcal{T}} \mid {T} \mmodels \neg S \}$,\\
& & ${\mathcal{T}} = {\mathcal{T}}_1 \cup {\mathcal{T}}_2,$ and ${\mathcal{T}}_1 \cap {\mathcal{T}}_2 = \emptyset.$
\end{tabular}
\end{equation}

\noindent Analogously to the static case, $C(T)$ indicates the true class of the multivariate time series $T$, and $\hat\varphi(T)$ indicates the class assigned by $\hat\varphi$.

\section{Temporal J48}

\noindent{\bf Entropy-based learning}. In ~\cite{DBLP:journals/ipl/HyafilR76}, it has been proved that computing the optimal decision tree is an NP-hard problem, where the notion of optimality is expressed as the relation between the height and the performance of the tree. In the perspective of practical applications of decision trees to real-life data, this justifies the use of algorithms that return sub-optimal trees, such as ID3~\cite{DBLP:journals/ml/Quinlan86}. ID3 is designed for static, categorical data, at it encompasses $k$-ary splits, but, without loss of generality, we focus our attention on binary splits only.

\medskip

ID3 is able to learn a tree from a purely categorical data set (i.e., $\bowtie \, \in \{=\}$ in Equation~\eqref{eq:decisions_static_dt}), and it uses the concepts of information gain and entropy to select the best decision at every node. By identifying frequencies with probabilities, one defines the {\em information conveyed} by $\mathcal D$ (or {\em entropy} of $\mathcal D$) by first measuring the amount of instances in $\mathcal D$ that belong to each class $C_i$ (let us denote this subset with $\mathcal D_{C_i}$), and, then, by computing:

\begin{equation}
Info(\mathcal{D}) = -\sum_{i=1}^{i=2}(\frac{|\mathcal D_{C_i}|}{|\mathcal{D}|}\log(\frac{|\mathcal D_{C_i}|}{|\mathcal{D}|})).
\end{equation}

\noindent Intuitively, the entropy is inversely proportional to the purity degree of $\mathcal{D}$ with respect to the class values. {\em Splitting}, which is the main {\em greedy} operation in learning a decision tree with ID3, is performed over a specific attribute $A$. When restricted to categorical attributes and binary splits, a split depends on a particular attribute $A$ and a particular value $a\in dom(A)$, which entail two subsets $\mathcal D_1$ and $\mathcal D_2$; the former (resp. the latter) contains all those instances $D$ such that $D\models (A=a)$ (resp., $D\models (A\neq a)$ - see Equation~\eqref{eq:sematincs_static_dt}). Thus, we can compute the {\em splitting information} of the pair $A,a$ as follows:

\begin{equation}\label{eq:info_cat}
InfoSplit(A,a,\mathcal{D}) = \sum_{i=1}^{i=2}\frac{|\mathcal{D}_i|}{|\mathcal{D}|}Info(\mathcal{D}_i),
\end{equation}

\noindent which implies that the {\em entropy of attribute} $A$ is defined as:

\begin{equation}\label{eq:entropy_attribute}
InfoAtt(A,\mathcal{D}) = \min_{a \in dom(A)} \{ InfoSplit(A,a,\mathcal{D}) \},
\end{equation}

\noindent and, finally, that the {\em information gain} of $A$ is defined as:

\begin{equation}\label{eq:info_gain}
Gain(A,\mathcal{D}) = Info(\mathcal{D}) - InfoAtt(A,\mathcal{D}).
\end{equation}

\noindent The algorithm ID3 is based on the idea of recursively splitting the data set over the attribute and the value of its domain that guarantee the greatest information gain, until a certain stopping criterion is met. When non-binary splits are allowed, the concept of splitting information must be slightly modified, but the underlying ideas remain. Given a temporal data set $\mathcal T$, one can use the abstraction algorithm presented in~\cite{DBLP:conf/iwinac/SciaviccoSV19} to obtain a discretized version of $\mathcal T$, in which every mutivariate time series becomes a timeline (as explained in Section~\ref{sect:preliminaries}). The algorithm {\em Temporal ID3}~\cite{DBLP:conf/jelia/BrunelloSS19} is able to extract a temporal decision tree that follows the general theory explained in the previous section using the same principles of entropy and information gain. As time series are represented in the form of timelines, Temporal ID3 takes decisions of the type $\langle X\rangle(A=a)$, where $A$ is a discretized attribute and $a$ is one of the possible propositions that emerged from the discretization, and establishes the interval relation, attribute, and propositional letter over which the split is performed according to the entropy principle. In other words, we have:

\begin{equation}\label{eq:info_temp_cat}
InfoSplit(A,X,a,\mathcal{T}) = \sum_{i=1}^{i=2}\frac{|\mathcal{T}_i|}{|\mathcal{D}|}Info(\mathcal{T}_i),
\end{equation}

\noindent where $X$ takes values in $\mathcal X\cup\{eq\}$, $eq$ being the interval temporal relation that captures the current interval only, and $T_1,T_2$ are computed as explained in the previous section (with $\alpha=1$ and $z=0$), and:

\begin{equation}\label{eq:entropy_temp_attribute}
InfoAtt(A,\mathcal{T}) = \min_{X\in\mathcal X\cup\{eq\},a \in dom(A)} \{ InfoSplit(A,X,a,\mathcal{T}) \}.
\end{equation}


\medskip

The algorithm C4.5~\cite{Quinlan93} is designed to allow ID3 to cope with numerical data. C4.5 uses exactly the same principles of entropy and information gain introduced for ID3. The main difference, in the static case, lies in allowing $\bowtie$ to take values in $\{\le,=\}$ in propositional decisions; indeed, if $A$ is numeric, the natural propositional decision is of the type $A\le a$:

\begin{equation}\label{eq:info_cat_c45}
InfoSplit(A,a,\bowtie,\mathcal{D}) = \sum_{i=1}^{i=2}\frac{|\mathcal{D}_i|}{|\mathcal{D}|}Info(\mathcal{D}_i),
\end{equation}

\noindent where $\bowtie \, \in\{\le,=\}$, and:

\begin{equation}\label{eq:entropy_attribute_c45}
InfoAtt(A,\mathcal{D}) = \min_{\bowtie\in\{\le,=\},a \in dom(A)} \{ InfoSplit(A,a,\bowtie,\mathcal{D}) \}.
\end{equation}

\noindent Incidentally, the obtained tree gives a new kind of information, that is, the values of the splitting points that give the most information. If we consider a temporal data set $\mathcal T$, in which multivariate time series are non-discretized, we obtain a similar result in the temporal case by simply allowing, as above, $\bowtie \, \in\{\le,=,>\}$; we may call the resulting algorithm {\em Temporal C4.5}. Observe that, as explained in the previous section, Temporal C4.5 has two new parameters, that is:

\begin{equation}\label{eq:info_temp_cat_c45}
InfoSplit(A,X,a,\bowtie,\alpha,z,\mathcal{T}) = \sum_{i=1}^{i=2}\frac{|\mathcal{T}_i|}{|\mathcal{D}|}Info(\mathcal{T}_i),
\end{equation}

\noindent where $X$ takes values in $\mathcal X\cup\{eq\}$, and $T_1,T_2$ are computed as explained in the previous section (but, this time, varying $\alpha\in(0,1]$ and $z\ge 0$), and

\begin{equation}\label{eq:entropy_temp_attribute_c45}
InfoAtt(A,\mathcal{T}) = \min_{\substack{X\in\mathcal X\cup\{eq\},\alpha\in(0,1],\\0\le z\le Max_z,a \in dom(A)}} \{ InfoSplit(A,X,a,\bowtie,\alpha,z,\mathcal{T}) \},
\end{equation}

\noindent where $Max_z$ must be fixed beforehand.

\medskip

\noindent{\bf A working implementation.} The open-source learning suite Weka~\cite{weka} offers the implementation the algorithm C4.5 in Java, called {\em J48}. The most important distinctive characteristic of one specific implementation over the others is the stopping condition; different stopping conditions may lead to different trees. J48 uses a very intuitive principle: having decided a minimal {\em purity degree} (i.e., a minimal value of $Info(\mathcal D)$, where $\mathcal D$ is associated to a leaf), the base case of the learner is fired on a node if its purity degree is high enough.

\medskip

Being object-oriented, such an implementation is ideal to test the predictive capabilities of the trees learned following the schema in Section~\ref{sec:theory} with minimal (yet, non-trivial) modifications. As a matter of fact, we can keep the entire structure of J48: model construction, stopping condition, and training/test performance indicator calculation and displaying. There are two main modifications required: \begin{inparaenum}[\it (i)] \item input data representation, and \item splitting management. \end{inparaenum} As far as the first point is concerned, we used an internal representation based on the string data type, that implicitly assumes that all temporal attributes have the same length and that there are no missing data. Strings have a simple internal structure, in which each value is separated from the next one by a semicolon. Splitting, on the other hand, is taken care by simply building the necessary Java classes that take care of the possible cases. {\em Temporal J48}, as we call the resulting implementation, requires the following parameters in addition to those already required by J48: \begin{inparaenum}[\it (i)] \item the value of $\alpha$ (which in this first experiment we did not optimized at each decision, unlike the general theory suggests); \item the value of $Max_z$; \item the reference intervals policy; \item the subset of the language HS that one allows during the learning phase. \end{inparaenum}

\section{Experiments and Results}\label{sect:experiments}

\begin{table}[t!]
\centering\footnotesize
\resizebox{0.9\textwidth}{!}{
\begin{tabular}{|c|c|c|c|c|c|}
\hline
\textbf{Dataset} & Train cases & Test cases & Channels & Length & Classes \\ \hline
\textbf{AtrialFibrillation  (AF)} & $24$ & $6$ & $2$ & $150$ & $3$ \\ \hline
\textbf{FingerMovements (FM)} & $104$ & $26$ & $28$ & $50$ & $2$ \\ \hline
\textbf{Libras (LI)} & $180$ & $45$ & $2$ & $45$ & $15$ \\ \hline
\textbf{LSST (LS)} & $168$ & $42$ & $6$ & $36$ & $14$ \\ \hline
\textbf{NATOPS (NA)} & $96$ & $24$ & $24$ & $51$ & $6$ \\ \hline
\textbf{RacketSports (RS)} & $96$ & $24$ & $6$ & $30$ & $4$ \\ \hline
\textbf{SelfRegulationSCP1 (S1)} & $96$ & $24$ & $6$ & $150$ & $2$ \\ \hline
\textbf{SelfRegulationSCP1 (S2)} & $96$ & $24$ & $7$ & $150$ & $2$ \\ \hline
\textbf{UWaveGestureLibrary (UW)} & $96$ & $24$ & $3$ & $150$ & $8$ \\ \hline
\end{tabular}
}
\caption{A summary of resampled datasets from~\cite{bagnall2018uea}.}
\label{tab:datasets}
\end{table}

\noindent{\bf Data sets.} In order to design a first systematic test aimed to establish the predictive capabilities of Temporal J48, we considered the public temporal data set from~\cite{bagnall2018uea}. From it, we have extracted nine data sets, which contain problems that vary from the medical context, to automatic recognition of sing language words, to classification of different racket sports based on the movements performed by the athletes. Some adaptations were necessary, taking into account two aspects: \begin{inparaenum}[\it (i)] \item the intrinsic computational inefficiency of Temporal J48 compared with existing methods, due to the substantial amount of information that can be extracted from its results, and \item the intrinsic unbalance between training and test cardinalities in the original settings in~\cite{bagnall2018uea}. \end{inparaenum} We modified the data sets as the result of several initial tests by: \begin{inparaenum}[\it (i)] \item trimming the number of temporal points for those data sets with too long time series, and, in particular, by limiting all time series to $N=150$, and \item re-sampling training and test instances to obtain a more standard $80\%-20\%$ ratio. \end{inparaenum} The resulting situation is summarized in Table~\ref{tab:datasets}.
%
%

\medskip

\begin{table}[t!]
\centering\footnotesize
\resizebox{0.9\textwidth}{!}{
\begin{tabular}{|c|c|c|c|c|c|c|c|c|c|}
\hline
\textbf{Dataset} & \textbf{AF} & \textbf{FM} & \textbf{LI} & \textbf{LS} & \textbf{NA} & \textbf{RS} & \textbf{S1} & \textbf{S2} & \textbf{UW} \\ \hline
J48 $1,0,0,0$      &    \underline{83.33}         &   \underline{50.00}          &  40.00           &  30.95           &  \underline{79.17}           &  70.83           &   \underline{66.67}          &  50.00           &   \underline{66.67}          \\
J48 $1,1,0,0$      &  \underline{83.33}           &  42.31           &   51.11          &   30.95          &  75.00           &   \underline{87.50}$^*$          &         \underline{66.67}    &   54.17          &   62.50          \\
J48 $1,1,1,1$      &    \underline{83.33}         &   42.31          &   \underline{64.44}          &   \underline{38.10}          &   62.50          &   79.17          &    \underline{66.67}         &   \underline{62.50}          &   54.17          \\ \hline
ED$_I$            &  83.33           &  \underline{76.92}           &  86.67           &  \underline{42.86}$^*$           &  70.83           &  79.17           &  66.67           &  \underline{66.67}           &  87.50           \\
DTW$_I$           &  \underline{100.00}$^*$          &  65.38           &  \underline{91.11}$^*$           &  33.33           &  \underline{87.50}$^*$           &  75.00           &    66.67         &  \underline{66.67}           &  91.67           \\
DTW$_D$           &   83.33          &   57.69          &  \underline{91.11}$^*$           &  40.48           &  \underline{87.50}$^*$           &  \underline{83.33}           &   \underline{83.33}$^*$         &  \underline{66.67}           &  \underline{95.83}$^*$           \\ \hline
T. J48 $0.5$       &   66.67          &  57.69           &  \underline{80.00}           &  23.81           &  \underline{83.33}           &  70.83           &  \underline{83.33}$^*$           &  54.17           &  62.50           \\
T. J48 $0.6$       &  66.67           &  57.69           &  71.11           &   \underline{26.19}          &  79.17           &  \underline{79.17}           &  66.67           &   \underline{75.00}$^*$          &  58.33           \\
T. J48 $0.7$       &    66.67         &   53.85          &   73.33          &   23.81         &  75.00           &   66.67          &    66.67         &   66.67          &   62.50          \\
T. J48 $0.8$       &   \underline{83.33}          &    \underline{80.77}$^*$         &    75.56         &   \underline{26.19}          &   75.00          &   62.50          &   66.67          &   62.50          &   \underline{66.67}          \\
T. J48 $0.9$       &  66.67           &   \underline{80.77}$^*$          &   71.11          &   23.81          &   66.67          &   62.50          &  66.67           &  70.83           &   \underline{66.67}          \\ \hline
\end{tabular}
}
\caption{Test results in terms of accuracy. Underlined results are the best ones in the group, and starred results are the absolute best ones.}\label{tab:results}
\end{table}

\noindent{\bf Experimental settings, results, and discussion.} We tested the effectiveness of Temporal J48 against feature-based and distance-based methods, in terms of test accuracy only. This is a highly limited comparison, as the major strength of our approach lies in the interpretability of the (temporal component of the) resulting model, and such a characteristics does not emerge from a purely numeric performance test. Yet, it is interesting to see how good Temporal J48 performs against non-interpretable methods. The results are summarized in Table~\ref{tab:results}: underlined results are the best ones for the category (feature-based, distance-based, or symbolic), and starred results are the absolute best ones. As for feature-based models, we considered the standard J48 executed on three combinations of abstractions of the temporal data set; for each channel or attribute we computed mean, standard deviation, skewness, and kurtosis, and we combined them in three different ways, each expressed in Table~\ref{tab:results} as a bit mask. So, for example, J48 with mask 1,1,0,0 means running the standard decision tree extraction algorithm on a abstracted data set with exactly two attributes per channel, namely mean and standard deviation. As for distance-based methods, we considered the standard, open-source available methods ED$_\mathit I$, DTW$_\mathit I$, and DTW$_\mathit D$, which require no parametrization. Finally, the parameter that we have used for Temporal J48 are: $0.5\le \alpha\le 0.9$, with 0.1 step, $Max_z=0$, and full HS.

\medskip

As it can be seen, different temporal data set are best dealt with different approaches. In five out nine cases distance-based methods behaved best; in one case feature-based behaved best, using, in particular, only mean and standard deviation. In all other cases, three, there was a run of Temporal J48 that performed better than every other method. As we have explained, obtaining the highest accuracy was {\em not} our initial motivation; nonetheless, comparing our method against the others in terms of (test) accuracy is a good proxy for its performance. Yet, as a matter of fact, our approach overcame our expectations: we obtained an interpretable classification model of each of the data sets, and in three cases our model was also the most performing one, indicating that our approach is worth pursuing further. In this first experiment we did not examined all possible parametrization that Temporal J48 offers; in particular, decisions were taken only on the 0-th derivative (so no trends, and no acceleration/decelerations of trends were taken into account), and the uncertainty value $\alpha$ was fixed at the same value for all intervals. This suggests that a further analysis of the predictive capabilities of Temporal J48 may result in even better performances. More importantly, in some of the cases, Temporal J48 obtained a nearly perfect classification of some of the classes (i.e., individual ROC curve close to 1); these cases give rise to temporal formulas (which can be read on the resulting tree) which, in a way, describe those classes from the temporal logic point of view.

\section{Conclusions}

In this paper we approached the problem of multivariate time series classification. Existing methods for classification of multivariate time series present good performances in terms of accuracy, but the extracted models are not interpretable, in particular in the temporal component. Distance-based methods are based on the concept of distance between series, and feature-based methods, while compatible with interpretable classifiers, flatten the temporal component of the data set. Based on a recently proposed algorithm (Temporal ID3), which is able to classify previously discretized multivariate time series, we developed Temporal C4.5 and realized its implementation, Temporal J48, following the same principle of describing time series using interval temporal logic. Temporal J48 is compatible with the well-known data mining suite Weka.

\medskip

The initial results show that the interval temporal logic HS is able to correctly describe the behaviour of multivariate time series. As a matter of fact, the predictive capabilities of Temporal J48 are comparable with those of existing methods, and superior to them in some cases, notwithstanding the fact that temporal decision tree models are interpretable even in the temporal component, and therefore undergo a very constrained learning phase (unlike non-interpretable methods, which are notoriously more adaptable). This suggests that temporal symbolic learning may be a promising topic, taking into account that, by examining the temporal component in an explicit way, several new learning parameters emerge that can be adjusted to improve the performances of the extracted models. Future developments of this line include, but are not limited to, exploring the predictive capabilities of variants of the language HS, and studying the possibility of adapting other well-known symbolic learning algorithms in the same way.

\bibliography{biblio}

\newpage

\section*{Appendix}

\begin{figure}[t!]
\centering\footnotesize
\begin{lstlisting}
<L> var5 <= -2.756591
|   <InvA> var5 <= 0.308951
|   |   <=> var2 > -0.916901
|   |   |   <InvB> var0 <= 2.832243: Badminton_Clear (6.0)
|   |   |   [InvB] var0 > 2.832243: Badminton_Smash (1.0)
|   |   [=] var2 <= -0.916901
|   |   |   <B> var3 <= -0.207743
|   |   |   |   <InvB> var0 > 4.115426
|   |   |   |   |   <D> var0 > 1.452113: Squash_ForehandBoast (3.0)
|   |   |   |   |   [D] var0 <= 1.452113: Squash_BackhandBoast (1.0)
|   |   |   |   [InvB] var0 <= 4.115426
|   |   |   |   |   <InvB> var0 <= -0.215688: Badminton_Smash (2.0)
|   |   |   |   |   [InvB] var0 > -0.215688: Badminton_Clear (3.0)
|   |   |   [B] var3 > -0.207743: Squash_ForehandBoast (14.0)
|   [InvA] var5 > 0.308951
|   |   <InvB> var5 <= -2.27452
|   |   |   <InvA> var0 <= -1.044682: Squash_BackhandBoast (3.0/1.0)
|   |   |   [InvA] var0 > -1.044682: Squash_ForehandBoast (7.0)
|   |   [InvB] var5 > -2.27452: Squash_BackhandBoast (21.0)
[L] var5 > -2.756591
|   <A> var0 <= 0.098773
|   |   <InvB> var0 > -0.960139
|   |   |   <B> var4 <= 0.625893: Badminton_Smash (16.0)
|   |   |   [B] var4 > 0.625893: Badminton_Clear (1.0)
|   |   [InvB] var0 <= -0.960139: Badminton_Clear (2.0)
|   [A] var0 > 0.098773
|   |   <L> var4 > 8.703901: Badminton_Smash (4.0)
|   |   [L] var4 <= 8.703901: Badminton_Clear (12.0)
\end{lstlisting}
\caption{One of the Temporal J48 models trained on data set RacketSports.}
\label{fig:output_temporal_j48}
\end{figure}

\noindent{\bf Time series classification methods.} We briefly review the literature of time series classification. We used some of the available techniques, described here, for a comparison against Temporal J48.

\medskip

Univariate time series classification is a well-studied problem in the literature; the reader can refer to~\cite{DBLP:journals/datamine/BagnallLBLK17} for an in-depth analysis on state-of-the-art methods for univariate time series classification. For the multivariate case, the classical accepted methods in the literature are {\em feature-based} or {\em distance-based}. Feature-based methods are very simple to understand as they are based on  extracting numerical or categorical descriptions from each channel. Such descriptions can be simple statistical values (e.g., mean, minimum, maximum, variance, skewness) or yes/no values that refer to the presence of certain patterns (e.g., shapelets). The collection of all descriptions can be then used as input to a classical, static learning algorithm~\cite{Wu2018AnOO}. In some cases, algorithms can be adapted to natively extract patterns from the temporal data sets, as it is the case of~\cite{DBLP:journals/computers/BrunelloMMS19}. The most widely accepted distance-based methods for multivariate time series classification is the classical {\em Nearest Neighbour} ($NN$) algorithm~\cite{cover} equipped with a proper notion of distance. In the univariate case, given two time series $T_1 = a_1,a_2,\ldots,a_N$ and $T_2=b_1,b_2,\ldots,b_N$, the {\em Euclidean distance (ED)} between $T_1$ and $T_2$ is simply the sum of the Euclidean distance between each pair $(a_i,b_i)$. The {\em dynamic time warping distance (DTW)} generalizes such concept by means of an alignment procedure that consists in constructing a $N \times N$ {\em distance matrix}, computed via dynamic programming, that allows one to find the alignment that minimizes the point-to-point Euclidian distance. In other words, DTW generalizes the notion of Euclidean distance from single points to single time series. In the multivariate case, in~\cite{SWK2015} this notion of distance is further generalized in two versions, named {\em DTW$_\mathit I$} ({\em independent} DTW) and {\em DTW$_\mathit D$} ({\em dependent} DTW), which differ by how the different channels are combined into a single distance. Thus, distance-based multivariate time series classification is traditionally solved via ED$_\mathit I$ (the independent multivariate generalization of the Euclidean distance), DTW$_\mathit I$, or DTW$_\mathit D$. Feature-based and distance-based methods present similar drawbacks: feature-based extract an explicit theory of a temporal data set, but such a theory is hardly interpretable, as it is written in the language of abstract indicators, and non-temporal, as the temporal component plays no role, while distance-based methods solve the classification problem in a black-box way, without extracting any symbolic theory at all. So, the methods of both groups are, in a way, non-interpretable. Finally, using Temporal ID3 (Section~\ref{sect:preliminaries}) to classify previously discretized multivariate time series can be thought of as a {\em timeline-based} classification method. In this particular taxonomy, Temporal C4.5 is a {\em symbolic} time series classification method.

\medskip

\begin{figure}[t]
\begin{tikzpicture}
\node(or1) at (0,0){
\begin{tabular}{ll}
$
T_1=\left\{
\begin{array}{lllllll}
A_1 &=& a_{1,1},&a_{1,2},&\ldots,&a_{1,N}\\
A_2 &=& a_{2,1},&a_{2,2},&\ldots,&a_{2,N} \\\
\ldots
\end{array}\right.
$
&
$C_1$
\\
$
T_2=\left\{
\begin{array}{lllllll}
A_1 &=& b_{1,1},&b_{1,2},&\ldots,&b_{1,N}\\
A_2 &=& b_{2,1},&b_{2,2},&\ldots,&b_{2,N} \\
\ldots
\end{array}\right.
$
&
$C_2$
\\
\ldots & \ldots \\
\end{tabular}
};
\node(or2) at (-4,-5) {
\begin{tabular}{cccc}
\hline
$A_1$ & $A_2$ & \ldots &  $C$ \\
\hline
$a_{1,1}$ & $a_{1,2}$ & \ldots &  $C_1$ \\
$a_{2,1}$ & $a_{2,2}$ & \ldots &  $C_1$ \\
 \ldots &  \ldots & \ldots   & $C_1$ \\
$b_{1,1}$ & $b_{1,2}$ & \ldots &  $C_2$ \\
$b_{2,1}$ & $b_{2,2}$ & \ldots &  $C_2$ \\
 \ldots &  \ldots & \ldots &    \ldots \\
 \ldots &  \ldots & \ldots &    \ldots \\
\hline
\end{tabular}
};
\node(or3) at (4,-5) {
\begin{tabular}{lllll}
\hline
$A_1$ & $A_2$ & \ldots &  $C$ \\
\hline
$a_{1,1};a_{1,2};\ldots$ & $a_{2,1};a_{2,2};\ldots$ & \ldots & $C_1$\\
$b_{1,1};b_{1,2};\ldots$ & $b_{2,1};b_{2,2};\ldots$ & \ldots & $C_2$\\
\ldots & \ldots  & \ldots & \ldots \\
\hline
\end{tabular}
};
\node(or2c) at (-2,0){};
\node(or2b) at (-1.8,-5){};
\node(or3b) at (0.5,-5){};
\draw[->](or1) -- (or2);
\draw[->](or2b) -- (or3b);
\end{tikzpicture}
\caption{Representation of a temporal data set: original data set (top), natural, tabular representation (bottom, left), and Temporal J48 internal representation (bottom, right).}\label{fig:datarep}
\end{figure}
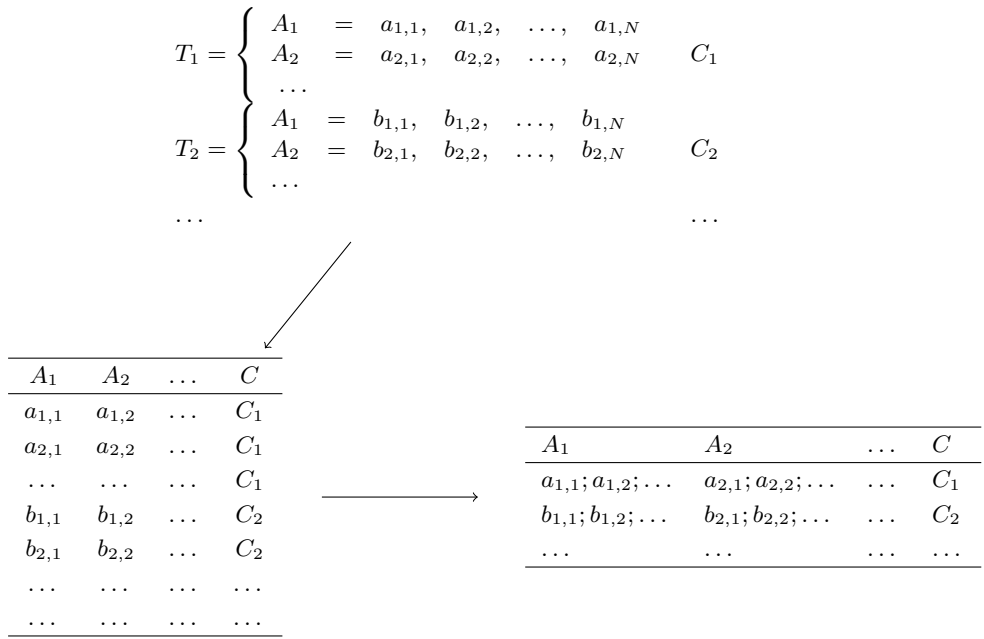

\noindent{\bf Temporal J48 internal representation.} In Figure~\ref{fig:datarep}, we see how data are represented in Temporal J48. Data are abstractly represented as at the top of the figure; the same data present naturally as a matrix, as at the bottom-left of the figure. Finally, we internally represent using a {\em string} data type, as in the bottom-right of the figure.

\medskip

\noindent{\bf A decision tree learned by Temporal J48.} Let analyze, more in depth, some of the results of our test with Temporal J48.

\medskip

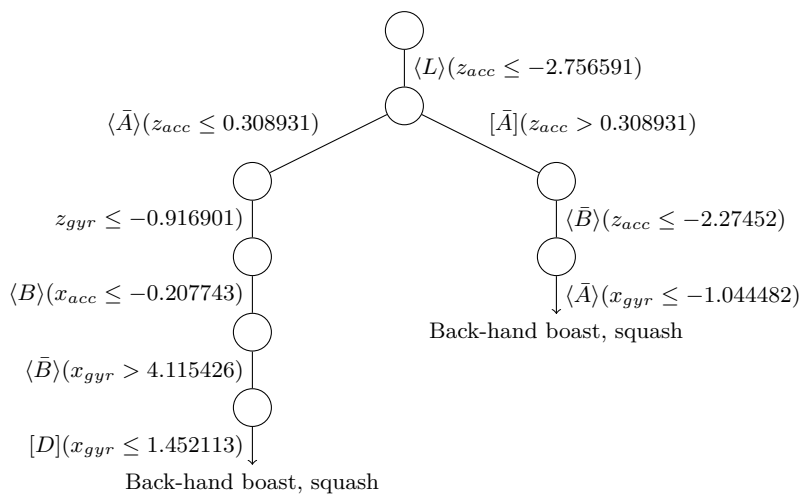
\begin{figure}[t!]
\centering\footnotesize
\begin{tikzpicture}[inner/.style={draw,circle,minimum size=0.5cm,minimum height=0.2cm},
							leaf/.style={draw,circle,minimum size=1cm}]

\node[inner] (root) at (0,0) {};
\node[inner] (0) at (0,-1) {};

\node[inner] (01) at (-2,-2) {};
\node[inner] (02) at (-2,-3) {};
\node[inner] (03) at (-2,-4) {};
\node[inner] (04) at (-2,-5) {};
\node (05) at (-2,-6) {Back-hand boast, squash};

\node[inner] (11) at (2,-2) {};
\node[inner] (12) at (2,-3) {};
\node (13) at (2,-4) {Back-hand boast, squash};

\draw (root) to [right] node {$\langle L \rangle (z_{acc} \leq -2.756591)$} (0);

\draw(0) to [above left] node {$\langle \bar{A} \rangle (z_{acc} \leq 0.308931)$} (01);
\draw(01) to [left] node {$z_{gyr} \leq -0.916901)$} (02);
\draw(02) to [left] node {$\langle B \rangle (x_{acc} \leq -0.207743)$} (03);
\draw(03) to [left] node {$\langle \bar{B} \rangle (x_{gyr} > 4.115426)$} (04);
\draw[->](04) to [left] node {$[D] (x_{gyr} \leq 1.452113)$} (05);

\draw (0) to [above right] node {$[\bar{A}] (z_{acc} > 0.308931)$} (11);
\draw (11) to [right] node {$\langle\bar{B}\rangle (z_{acc} \leq -2.27452)$} (12);
\draw[->] (12) to [right] node {$\langle\bar{A}\rangle (x_{gyr} \leq -1.044482)$} (13);
\end{tikzpicture}
\caption{Extracted theory for the movement of back-hand boast during squash.}
\label{fig:theory_squash_backhandboast}
\end{figure}

Consider the temporal data set RacketSports (see Section~\ref{sect:experiments}), in which each multivariate time series describes the movements of an athlete playing badminton or squash whilst wearing a smart watch, which relayed the $X,Y,Z$ coordinates for both a gyroscope and an accelerometer to a smart phone. More in particular, four classes are identified, two movements during the activity of playing squash, that is, {\em back-hand boast} and {\em fore-hand boast}, and two movement during the activity of playing badminton, that is, {\em smash} and {\em clear}. These movements are described by six channels, which contain the values of the sensors attached for each physical dimension at each moment of time. These variables are codified as follows: $Var_0,Var_1,$ and $Var_2$ (resp., $Var_3,Var_4,$ and $Var_5$) are the gyroscope (resp., accelerometer) values for $X,Y$ and $Z$. Run with $\alpha=0.6$ on this data set, Temporal J48 returned the tree in Figure~\ref{fig:output_temporal_j48}, which has, in test phase, the performances shown in Tab.~\ref{tab_perf}. By focusing on the squash back-hand boast movement only, which has a 0.94 ROC area, one can extract from Figure~\ref{fig:output_temporal_j48} a formula of HS, shown in Figure~\ref{fig:theory_squash_backhandboast}, that describes such a movement along the temporal component. This proves that our model extraction method allows one to {\em interpret} the underlying theory. In opposition, feature-based methods flatten the temporal component, so while a symbolic theory is extracted, it is not temporal, and distance-based methods do not extract a theory at all. 

\begin{table}[t!]
\centering\footnotesize
\resizebox{0.9\textwidth}{!}{
\begin{tabular}{|c|c|c|c|c|c|c|c|c|}
\hline
{\bf TP} &  {\bf FP}  &  {\bf Prec.} &  {\bf Rec.}  & {\bf F-M}  & {\bf MCC}  &   {\bf ROC} &  {\bf PRC} &   {\bf Class}\\
\hline
0.66 &   0.05 &   0.80  &    0.66 &   0.72 &      0.65 &    0.80 &    0.61 &  Smash, badminton \\
\hline
0.83 &   0.11 &   0.71 &     0.83 &   0.76 &     0.68 &   0.86 &    0.63 &    Clear, badminton \\
\hline
0.66 &   0.00 &   1.00 &     0.66 &   0.80 &     0.77 &   0.83 &    0.75 &    Fore-hand boast, squash\\
\hline
1.00 &   0.11 &   0.75 &     1.00 &   0.85 &     0.81 &   0.94 &    0.75 &    Back-hand boast, squash\\
\hline
\end{tabular}
}\caption{Test performances for the RacketSports data set.}\label{tab_perf}
\end{table}

\end{document}